%% file: acl_latex.tex
\documentclass[11pt]{article}

\usepackage[preprint]{acl}

\usepackage{times}
\usepackage{latexsym}

\usepackage[T1]{fontenc}

\usepackage[utf8]{inputenc}

\usepackage{microtype}

\usepackage{inconsolata}

\usepackage{graphicx}

\usepackage{booktabs} 
\usepackage{graphicx} 
\usepackage{xcolor} 
\usepackage{cite} 
\usepackage{pifont}
\usepackage{hyperref}
\usepackage{amsmath}

\usepackage{amsfonts}

\usepackage{makecell}


\definecolor{darkgreen}{RGB}{59,125,35}
\definecolor{darkred}{RGB}{139,0,0}

\newcommand{\correctmark}{\textcolor{darkgreen}{\ding{51}}} 
\newcommand{\errormark}{\textcolor{darkred}{\ding{55}}} 
\usepackage{tabularx}
\usepackage{adjustbox}

\usepackage{multirow} 
\title{FinBoardBench: Benchmarking Dynamic Wealth Management and Strategic Financial Reasoning of LLMs via Board Game Simulations}

\author{
  Xuesi Hu$^{1,\,2,\,*}$~~
  Peng Wang$^{1,\,3,}$\thanks{~~Equal Contribution.}~~
  Jinpeng Miao$^{3}$~~
  Xilin Tao$^{1}$~~
  Caiwei Li$^{4}$~~
  \\
  \textbf{Yue Ma$^{1}$~~
  Jie He$^{1}$~~
  Qiancheng Zhang$^{2}$~~
  Yuntao Zou$^{5}$~~
  Dagang Li$^{1,\,3,}$}\thanks{~~Corresponding Author.}
  \\
	$^{1}$~School of Computer Science and Engineering, Macau University of Science and Technology, Macau, China \\
  $^{2}$~School of Economics, Anhui University, Anhui, China\\
	$^{3}$~SKLPlanets, Macau University of Science and Technology, Macau, China\\
  $^{4}$~Department of Computer and Information Science, University of Macau, Macau, China\\
  $^{5}$~School of Energy and Power Engineering, Huazhong University of Science and Technology, Hubei, China\\
	\texttt{xuesihu13@gmail.com}, \texttt{wpengxss@gmail.com}, \texttt{dgli@must.edu.mo}\\
}

\usepackage[most]{tcolorbox}

\tcbset{
  promptbox/.style={
    width=\linewidth,
    breakable,
    enhanced,
    colback=blue!4!white,
    colframe=black!75,
    colbacktitle=black!85,
    coltitle=white,
    fonttitle=\bfseries\footnotesize,
    fontupper=\footnotesize,
    boxrule=0.5pt,
    arc=1pt,
    left=4pt,
    right=4pt,
    top=6pt,
    bottom=4pt,
    attach boxed title to top left={yshift=-0.08in,xshift=0.10in},
    boxed title style={boxrule=0pt,colframe=white},
  }
}
\newtcolorbox{PromptBox}[2][]{promptbox,title=#2,#1}

\begin{document}
\maketitle
\begin{abstract}
  Recently, large language models (LLMs) have achieved superior performance in static financial reasoning and simple dynamic trading tasks. However, existing static financial benchmarks are insufficient to assess the dynamic wealth management and financial decision-making capabilities of LLMs in real-world environments. To bridge this gap, we present FinBoardBench, an evaluation suite based on three classic financial board games: Cashflow, Acquire, and Monopoly. FinBoardBench assesses a comprehensive set of financial skills, including personal cash flow management with debt balancing, corporate investment and acquisition forecasting, and competitive trade negotiations with asset auctions. Our experiments with 9 advanced LLMs reveal that while exhibiting basic long-term planning and investment logic, they fail to effectively leverage complex interactions for profit, and their strong static reasoning performance does not transform into successful dynamic decision-making. Notably, they tend to prioritize immediate asset acquisition over maintaining sufficient liquidity, making them vulnerable to financial crises triggered by random events. We hope that FinBoardBench can provide a valuable reference for more intelligent LLM-based decision-making systems in the future.
\end{abstract}

\section{Introduction}


Large language models (LLMs) have demonstrated advanced performance across natural language processing tasks~\citep{10.1145/3641289,wang2025large,khan2025bridging}, prompting exploration of their potential in financial applications~\citep{ragavi2025llms,fu2025new}. Prior studies have proposed a variety of financial benchmarks to evaluate LLMs' abilities to process structured data and perform specific financial tasks~\citep{10.1145/3604237.3626869,dong2025large}. Notable examples include PIXIU~\citep{xie2023pixiu}, FinBen~\citep{NEURIPS2024_adb1d9fa}, and FinMaster~\citep{jiang2025finmaster}. These benchmarks effectively assess static reasoning, such as predicting stock trends, evaluating accounting principles, or answering knowledge-based questions.

However, financial decision-making in the real world is inherently dynamic and multidimensional. Models not only reason over available information but also adapt to stochastic events, manage liquidity under uncertainty, maintain long-term solvency, and strategically interact with competing agents~\citep{doi:10.1126/science.ade9097,valmeekam2023planbench,fu2023improving}. Existing static benchmarks primarily evaluate the financial knowledge and reasoning ability of LLMs, but hardly capture decision-making performance in complex environments where dynamic and multiple mechanisms coexist. Specifically, (1) how an agent balances short-term liquidity with long-term investments; (2) how it effectively utilizes various rules and actions across hundreds of decisions to place the game; (3) how it simultaneously competes with other agents and copes with stochastic financial events. While some research has explored the performance of LLMs in actual~\citep{10638546,chen2025stockbench} and simulated financial markets~\citep{lopez2025can,li-etal-2025-investorbench}, the dynamic and massive nature of real-world data makes it difficult to abstract into a unified benchmark for evaluating real financial decision-making capabilities. Consequently, the ability of LLMs to navigate complex, dynamic financial environments remains underexplored.

\input{table/intro_.tex}

Recent work has introduced game-based environments to fully evaluate the capabilities of LLMs in dynamic environments~\citep{NEURIPS2024_31911709,wang2025can,10752360}, and has achieved significant success. Inspired by these approaches, we propose FinBoardBench, a novel dynamic evaluation suite based on three classic financial board games: Cashflow, Acquire, and Monopoly. Financial games not only provide rigorous, controlled environments with clear victory criteria and transparent states, but also evaluate a broad spectrum of financial skills essential for real-world decision-making~\citep{CANNISTRA2024825}. These three games target complementary financial competencies: ($i$) \textbf{Cashflow} evaluates cash flow management by requiring players to balance debt and passive income; ($ii$) \textbf{Acquire} assesses corporate investment forecasting, including strategic asset acquisitions and mergers; ($iii$) \textbf{Monopoly} focuses on competitive resource allocation, open trade negotiations, and asset auctions. Compared to existing benchmarks (see Table~\ref{tab:compare}), FinBoardBench emphasizes dynamic, multi-step decision-making. Rather than simply answering financial questions, LLMs must manage multi-asset portfolios and survive stochastic risks in these interactive environments.

We conduct comprehensive experiments in FinBoardBench, including 9 advanced LLMs: GPT-5.4, Gemini-3.1-Pro Preview, GLM-5.1, HY3 Preview, DeepSeek V4 Pro, Doubao-2.0-Pro, Kimi K2.6, Qwen-3.6-Plus, and Mimo V2.5 Pro. Across the three games, our evaluation covers 108 game matches, comprising 4,282 rounds and 15,602 active player turns. Our results show that: ($i$) LLMs exhibit basic long-term planning and management logic, enabling them to play financial games, but lack the awareness and ability to use complex interactions to generate profits. ($ii$) LLMs prioritize immediate acquisition of assets rather than maintaining sufficient liquidity, making them vulnerable to financial crises caused by random events. ($iii$) The advanced LLMs perform well on static financial benchmarks, but the capability does not transform into effective financial decision-making ability, which leads to failure in dynamic financial games.

In summary, our contributions are as follows:

\begin{itemize}
  \item To the best of our knowledge, we are the first to consider evaluating the wealth management and financial decision-making capabilities of LLMs in game-based dynamic environments.
  \item We introduce FinBoardBench, a dynamic financial benchmark including multiple financial games, which provides a controlled environment for exploring the financial decision-making of LLMs.
  \item We conduct extensive experiments demonstrating that LLMs can handle dynamic financial events but lack the ability to execute complex actions and poorly manage cash flow.
\end{itemize}

\begin{figure*}[t!]
  \begin{center}
    \centerline{\includegraphics[width=\textwidth]{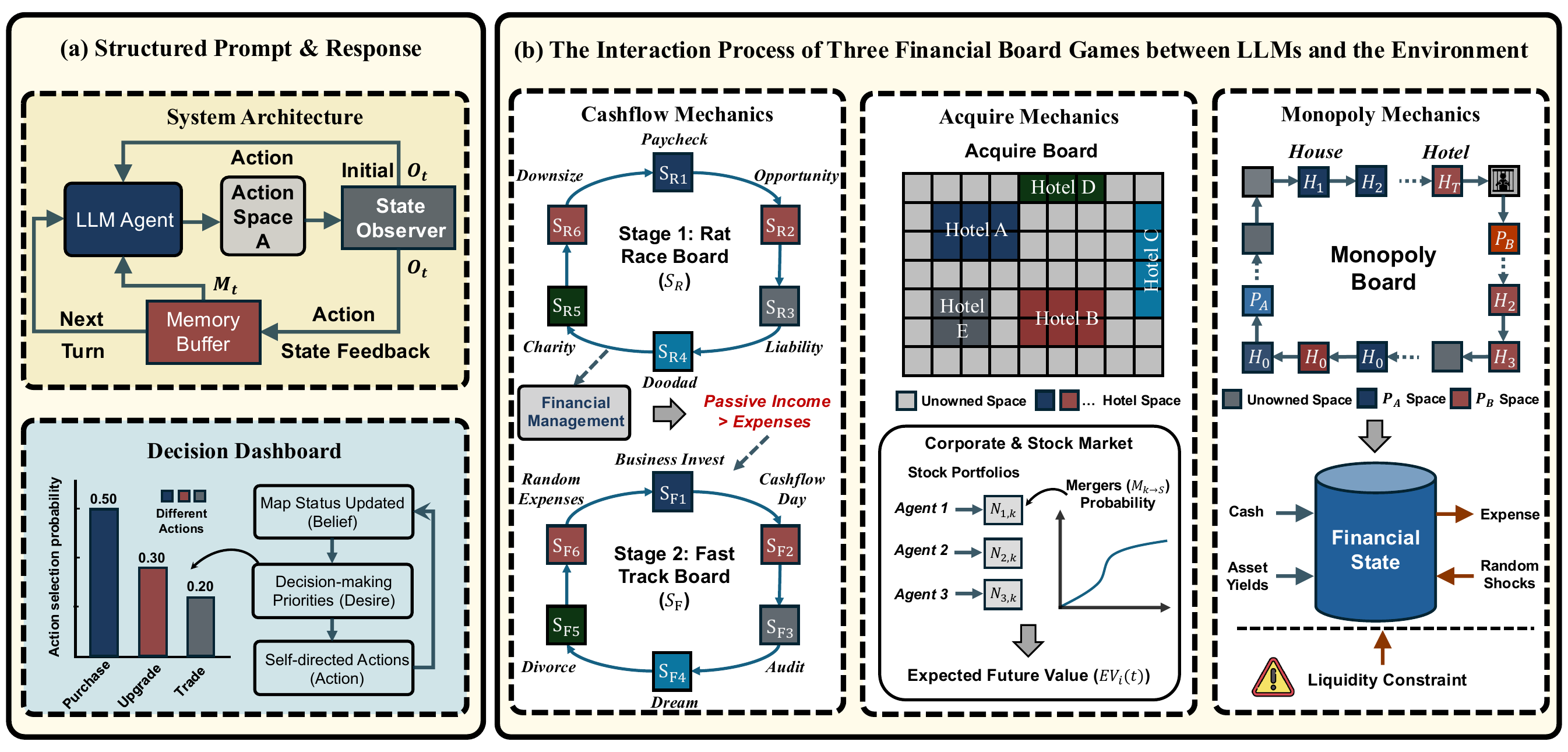}}
    \caption{The overall framework of FinBoardBench. (a) LLM-game interaction: System Architecture represents LLMs getting their current state and past actions to make decisions after moving, and then immediately updating with new information; Decision Dashboard represents LLMs weighing all available actions and choosing one to execute. (b) Game mechanics: In Cashflow, players achieve financial freedom that leaves the rat race for the fast track to reach their dream. In Acquire, players place tiles to found and merge hotels, then buy stocks to accumulate the greatest wealth. In Monopoly, players move around the board, buy property, and try to monopolize the real estate to win.}
    \label{Fig:method}
  \end{center}
\end{figure*}

\section{FinBoardBench}
We use three financial board games to assess LLMs' financial decision-making capabilities in a dynamic financial environment. This section details the construction of FinBoardBench, including the game background(\S~\ref{sec:Game_Background}) and the overall framework(\S~\ref{sec:Overall_Framework}).

\subsection{Game Background}
\label{sec:Game_Background}
Cashflow\footnote{\url{https://en.wikipedia.org/wiki/Cashflow_101}} is an educational game that simulates financial and economic aspects of daily life to improve people's financial literacy. It involves investing in real estate, businesses, and financial products to achieve financial freedom. Acquire\footnote{\url{https://en.wikipedia.org/wiki/Acquire}} is a financial strategy game in which players place building tiles and merge hotel chains. Each time a hotel is merged, the major shareholders receive a profit, and the player who ultimately earns the most profit wins. Monopoly\footnote{\url{https://en.wikipedia.org/wiki/Monopoly_(game)}} is an incorporated financial elements game, including market monopolies, wealth distribution, and resource control. Players buy properties, build monopolies, and manage limited cash to deal with random events like paying rent. These three games are classic financial board games that effectively evaluate LLMs' real-time decision-making and long-term planning abilities in a dynamic financial environment with multi-agent interaction.

\subsection{Overall Framework}
\label{sec:Overall_Framework}
This section outlines the complete process for gameplay. The specific framework is illustrated in Figure~\ref{Fig:method}.

\noindent

\textbf{Environment Construction.}
According to Wikipedia and other sources, we use the classic game board with rules closely aligned to actual offline game mechanics. Detailed game information is provided in Appendix~\ref{sec:cashflow_game_rule_appendix},~\ref{sec:acquire_game_rule_appendix}, and
~\ref{sec:monopoly_game_rule_appendix}. Additionally, following the mainstream game benchmark setting~\citep{NEURIPS2024_31911709,Fan_Chen_Jin_He_2024,lin-etal-2025-gamebot}, we input the global game information to LLMs, such as the current board situation and the states of all players. To better evaluate the financial decision-making capabilities of LLMs in dynamic environments, we provide each player with abundant context round by round, including passive events and active decisions from the previous rounds and the current round. Specifically, the structure for each player is as follows:

\textbf{Step 1}: Assume that the player moves on the board by rolling two dice, generating a random variable $d_t = d_{t,1} + d_{t,2}$, where $d_{t,i} \sim \mathrm{Unif}\{1,\dots,6\}$. The position of player $pos_t$ is updated as:
\begin{equation}
pos_t = (pos_{t-1} + d_t) \bmod N.
\end{equation}

\textbf{Step 2}: After reaching position $pos_t$, the player handles landing events, such as paying taxes. The update of game status can be defined as follows:
\begin{equation}
O_t^{(1)} = \mathcal{T}_{\mathrm{game}}(O_t^{(0)}, pos_t),
\end{equation}
where $O_t^{(0)}$ represents the initial state observation before reaching the new position; $\mathcal{T}_{\mathrm{game}}$ represents a game-driven function that updates the game state without player intervention.

\textbf{Step 3}: Next, following rational-player theory~\citep{Zagare1984-ZAGGTC,osborne1994course}, we decompose each decision into belief, desire, and action. The LLM loops through multiple autonomous sub-decisions $k = 1,2,\dots,K_t$ until reaching the maximum decision limit $K_{\max}$ or ending the turn. During each sub-step $k$, the execution process of LLMs is as follows:

(1) Beliefs: The LLM obtains the current observation $O_t^{(k)}$ and historical memory $M_t$. The belief state $\mathcal{B}_{t,k}$ is defined as follows:
\begin{equation}
\mathcal{B}_{t,k} = f_{\mathrm{belief}}(O_t^{(k)}, M_t),
\end{equation}
where $f_{\mathrm{belief}}$ represents the mapping function.

(2) Desire: Based on the current belief state $\mathcal{B}_{t,k}$, the LLM assigns a priority to each available action:
\begin{equation}
D_{t,k}(a) = f_{\mathrm{desire}}(a, \mathcal{B}_{t,k}), \quad a \in \mathcal{A}_{t,k}.
\end{equation}

(3) Action: The LLM selects a specific action from the currently available action list $\mathcal{A}_{t,k}$:
\begin{equation}
a_{t,k} = \arg\max_{a \in \mathcal{A}_{t,k}} D_{t,k}(a),
\end{equation}
where $a_{t,k}$ represents the chosen action and $D_{t,k}(\cdot)$ represents the desire function that assigns a priority value to available actions.

The $a_{t,k}$ updates the state through the player-driven transition function $\mathcal{T}_{\mathrm{player}}$:
\begin{equation}
O_t^{(k+1)} = \mathcal{T}_{\mathrm{player}}(O_t^{(k)}, a_{t,k}),
\end{equation}

\textbf{Step 4}: When $a_{t,k} = \text{"End Turn"}$ or $k$ reaches the maximum decision limit $K_{\max}$, the turn ends. The system records the player's final state and the events that occurred during the turn, marking each event as either passive or active $E_t = \{e_{t,1}, \dots, e_{t,m}\}$, $A_t = \{a_{t,1}, \dots, a_{t,K_t}\}$. The Memory Buffer is updated via the feedback loop:
\begin{equation}
M_{t+1} = M_t \cup \{(O_t^{\mathrm{final}}, E_t, A_t)\}.
\end{equation}

\noindent
\textbf{Cashflow.} The Cashflow represents a dynamic optimization game. The player must balance the long-term goal of achieving financial freedom with the short-term events to cope with random liquidity shocks. The dynamic cash flow equation is:
\begin{equation}
C_{t+1} = C_t + \Pi_t - \epsilon_t(\omega),
\end{equation}
where $C_t$ and $\mathcal{P}_t$ represent the player's cash and portfolio at time $t$, respectively; $X_t$ represents the player's financial decision; $I_{\mathrm{base}}$ is the base salary; $Y_t := \sum_{p \in \mathcal{P}_t} y(p)$ represents passive income; $R_t$ represents total expenses; $\Delta C(X_t)$ represents the net cash change induced by decision $X_t$; $\Pi_t := I_{\mathrm{base}} + Y_t - R_t + \Delta C(X_t)$ denotes the net cash flow; and $\epsilon_t(\omega)$ represents random shocks drawn from a probability distribution.

The objective of the player is to minimize the expected time to financial freedom $\tau$, formulated as $ \min_{\{X_t\}_{t \geq 0}} \mathbb{E}[\tau]$ subject to: $ Y_{\tau} \geq R_{\tau}$, and the liquidity constraint: $\mathbb{P}(C_t < 0) \leq \delta, \quad \forall t < \tau$.

\noindent
\textbf{Acquire.} The Acquire game simulates corporate mergers and acquisitions, which assesses the ability of multiple players to predict asset growth and hedge risks. Unlike static reasoning tasks, stock values in Acquire are uncertain and depend on the placement of tiles on the board. The expected future portfolio value $\mathrm{EV}_i(t)$ for player $i$ is:
\begin{equation}
  \mathrm{EV}_i(t) = C_{i,t} + \sum_{k \in \mathcal{K}_t} N_{i,k} \cdot \Psi_k(t) + \Omega_i(t),
\end{equation}

where $C_{i,t}$ is limited capital; $B_t$ denotes the board state; $\mathcal{K}_t$ and $\mathcal{S}_t$ denote active and potential acquiring corporations, respectively; $N_{i,k}$ represents shares in corporation $k$; $\Psi_k(t)=\mathbb{E}[V_k(B_{t+n})\mid B_t]$ is the expected future stock value; and $\Omega_i(t)=\sum_{k\in\mathcal{K}_t}\sum_{S\in\mathcal{S}_t}\mathbb{P}(M_{k\to S}\mid B_t)\Gamma_i(k,S,t)$ is expected merger premium, where $\Gamma_i(k,S,t)$ denotes the merger bonus from $k$ merging into $S$.

\begin{figure*}[t!]
  \begin{center}
    \centerline{\includegraphics[width=\textwidth]{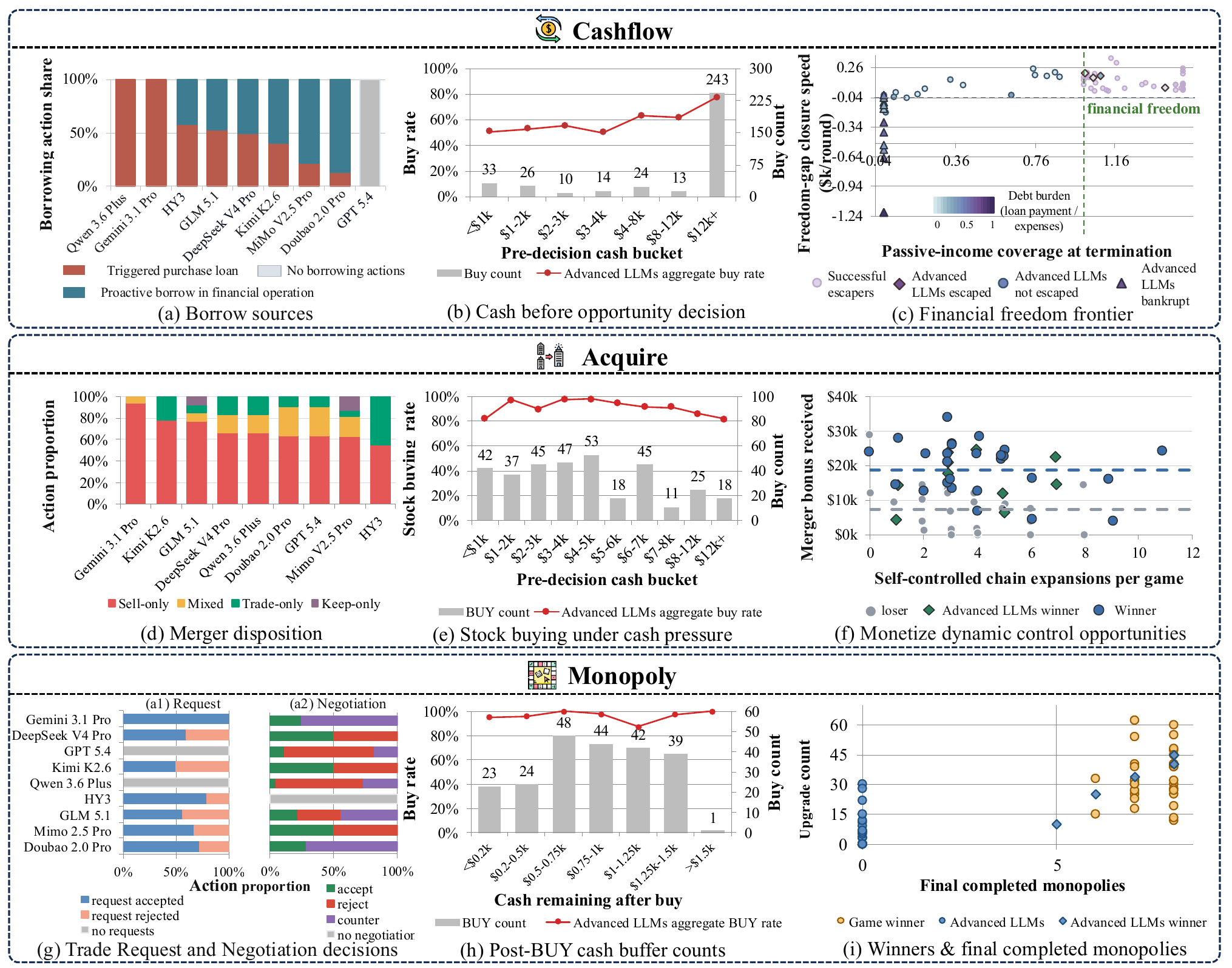}}
    \caption{Statistical analysis of LLMs' performance across three games.}
    \label{Fig:main_results}
  \end{center}
\end{figure*}

\textbf{Monopoly.} The Monopoly game constructs a multi-agent competitive environment with complex social interactions such as transactions and auctions. It transcends simple asset pricing and tests whether players can accurately infer the value of assets. The player's utility function is as follows:
\begin{equation}
U_i = V(Assets_i) - \lambda \cdot \max \left(0, \mathbb{E}[L_i] - C_i \right)^2,
\end{equation}
where $\pi(s)$ represents the probability of landing on space $s$; $\mathbb{E}[L_i]$ represents player $i$'s expected rental expenditure in the next $k$ rounds, calculated as $k \sum_{s \in Opponent} \pi(s) \cdot \mathrm{Rent}(s)$; $C_i$ denotes player $i$'s current cash; $\lambda$ represents the risk aversion to bankruptcy; $V(Assets_i)$ is the expected present value of the asset of player.

The goal of transaction $\tau$ between player $i$ and player $j$ is to optimize resource allocation through negotiation:
\begin{equation}
U_i(S_{t+1}^{\tau}) > U_i(S_t) \land U_j(S_{t+1}^{\tau}) > U_j(S_t),
\end{equation}
where $S_t$ represents the pre-trade state; $S_{t+1}^{\tau}$ represents the post-trade state. 

The key dynamic constraint is bankruptcy. It occurs when a player's stochastic liability $L_t$ exceeds their maximum liquidatable net worth.

\section{Experiments}
\subsection{Experimental Setting.} 
We evaluate 9 advanced LLMs in FinBoardBench: GPT-5.4, Gemini-3.1-Pro Preview, GLM-5.1, HY3 Preview, DeepSeek V4 Pro, Doubao-2.0-Pro, Kimi K2.6, Qwen-3.6-Plus, and Mimo V2.5 Pro. All LLMs use non-thinking mode and default settings in the API provider. To improve the interpretability of the ability comparison, reduce the effect of luck~\citep{doi:10.1126/sciadv.adn2654}, and focus more on the key actions of advanced LLMs. For each game match, we place one advanced LLM as a player and match it against three strong LLMs: DeepSeek V4 Flash, Qwen-3.5-397B-A17B, and Gemini-3.1-Flash-Lite-Preview. The starting positions of the advanced LLM are rotated across positions 1 through 4, while the others fill the remaining positions randomly. Across the three games, our evaluation covers 108 game matches, comprising 4,282 rounds and 15,602 active player turns.

\subsection{Results} 
Our findings are as follows:

\textit{\textbf{(1) LLMs exhibit passive decision-making in the dynamic financial environment.}} In Cashflow, as shown in Figure~\ref{Fig:main_results} (a), over 50\% originated from borrowing triggered by failed purchases among the loans obtained by the most advanced LLMs. This indicates that their financing behavior is often induced by failed purchase than proactive management. In Acquire, as shown in Figure~\ref{Fig:main_results} (d), merger disposition is dominated by sell-only actions (>65\% on average) and almost no keep-only choices, indicating preference for immediate cash over future earnings. In Monopoly, as shown in Figure~\ref{Fig:main_results} (g), among 86 completed deals initiated by advanced LLMs, 82 finish within one round; across both initiator and receiver, only 33.3\% negotiation decisions choose counter, showing a preference to accept or reject rather than bargain. Overall, advanced LLMs often react to visible state changes but fail to fully use strategic interaction mechanisms.

\textit{\textbf{(2) LLMs focus on asset acquisition but often overlook liquidity.}} In Cashflow, As shown in Figure~\ref{Fig:main_results} (b), Advanced LLMs still buy in 50.8\% of opportunities with cash below \$1k, and the aggregate buy rate remains relatively stable at 50.0\%-63.2\% across cash buckets below \$12k, indicating weak liquidity-sensitive purchasing. In acquire, as shown in Figure~\ref{Fig:main_results} (e), the purchase frequency of Advanced LLMs consistently exceeds 80\% and still buys with 89.0\% probability holding cash below \$1k, showing weak capacity for budget control. In Monopoly, as shown in Figure~\ref{Fig:main_results} (h), even if LLMs lack liquidity, they still choose to buy real estate. Specifically, the cash below \$500 after 47 purchases and 23 times below \$200. Overall, advanced LLMs keep buying even under low-cash conditions, overvaluing asset accumulation while underestimating liquidity preservation.

\textit{\textbf{(3) LLMs are unable to transform their financial knowledge into effective decision-making capabilities, leading to struggles in dynamic environments.}} In Cashflow, only 11.11\% of advanced LLMs run escape, while 44.44\% end in bankruptcy, and 44.44\% are terminated after an opponent wins first. As shown in Figure~\ref{Fig:main_results} (c), failed advanced LLMs remain far below the financial freedom threshold, with average passive-income coverage of only 0.19. In Acquire, advanced LLMs win at a rate of only 27.77\%. As shown in Figure~\ref{Fig:main_results} (f), despite comparable control opportunities, their losses earn only \$7.4k in merger bonuses versus \$18.6k for winners, revealing dynamic monetization failures. In Monopoly, as shown in Figure~\ref{Fig:main_results} (i), advanced LLMs win at a rate of only 13.88\% and mostly remain in the low monopoly control, low upgrade region. This indicates that they fail to effectively acquire assets to establish property group monopolies to pressure their opponents. These poor performances demonstrate that advanced LLMs struggle to perform effectively in dynamic financial environments.

\section{Analysis}

\begin{table}[t]
\centering
\resizebox{\columnwidth}{!}{%
\begin{tabular}{llcc}
\toprule
Game & Model & Non-thinking & Thinking (Enabled / Max) \\
\midrule
\multirow{3}{*}{\textbf{Cashflow}} & DeepSeek V4 Pro & $0/4$ (0.0\%) & $2/4$ (50.0\%) \\
& Mimo V2.5 Pro & $0/4$ (0.0\%) & $1/4$ (25.0\%) \\
& Qwen 3.6 Plus & $0/4$ (0.0\%) & $0/4$ (0.0\%) \\
\midrule
\multirow{3}{*}{\textbf{Acquire}} & DeepSeek V4 Pro & $2/4$ (50.0\%) & $2/4$ (50.0\%) \\
& Mimo V2.5 Pro & $1/4$ (25.0\%) & $1/4$ (25.0\%) \\
& Qwen 3.6 Plus & $1/4$ (25.0\%) & $2/4$ (50.0\%) \\
\midrule
\multirow{3}{*}{\textbf{Monopoly}} & DeepSeek V4 Pro & $0/4$ (0.0\%) & $1/4$ (25.0\%) \\
& Mimo V2.5 Pro & $0/4$ (0.0\%) & $1/4$ (25.0\%) \\
& Qwen 3.6 Plus & $0/4$ (0.0\%) & $0/4$ (0.0\%) \\
\bottomrule
\end{tabular}
}
\caption{Win-rate comparison between non-thinking and maximum-thinking modes.}
\label{tab:thinking_mode_winrate}
\end{table}

\subsection{A deeper level of thinking model in LLMs will not lead to significant improvement}
Since current LLMs possess hybrid thinking capabilities, and longer thinking depths typically lead to stronger performance, we evaluated the financial decision-making capabilities of current models in their highest thinking mode. Using the same experimental setup as the main experiment and the highest thinking mode, we evaluated the performance of DeepSeek V4 Pro, Mimo V2.5 Pro, and Qwen 3.6 Plus on FinBoardBench. As shown in Table~\ref{tab:thinking_mode_winrate}, increasing the depth of thinking has a limited effect on improving the win rate in LLMs. This indicates that current LLMs, even with deeper thinking, cannot significantly improve the model's ability in dynamic financial environments.

\begin{figure}[t]
  \centerline{\includegraphics[width=\columnwidth]{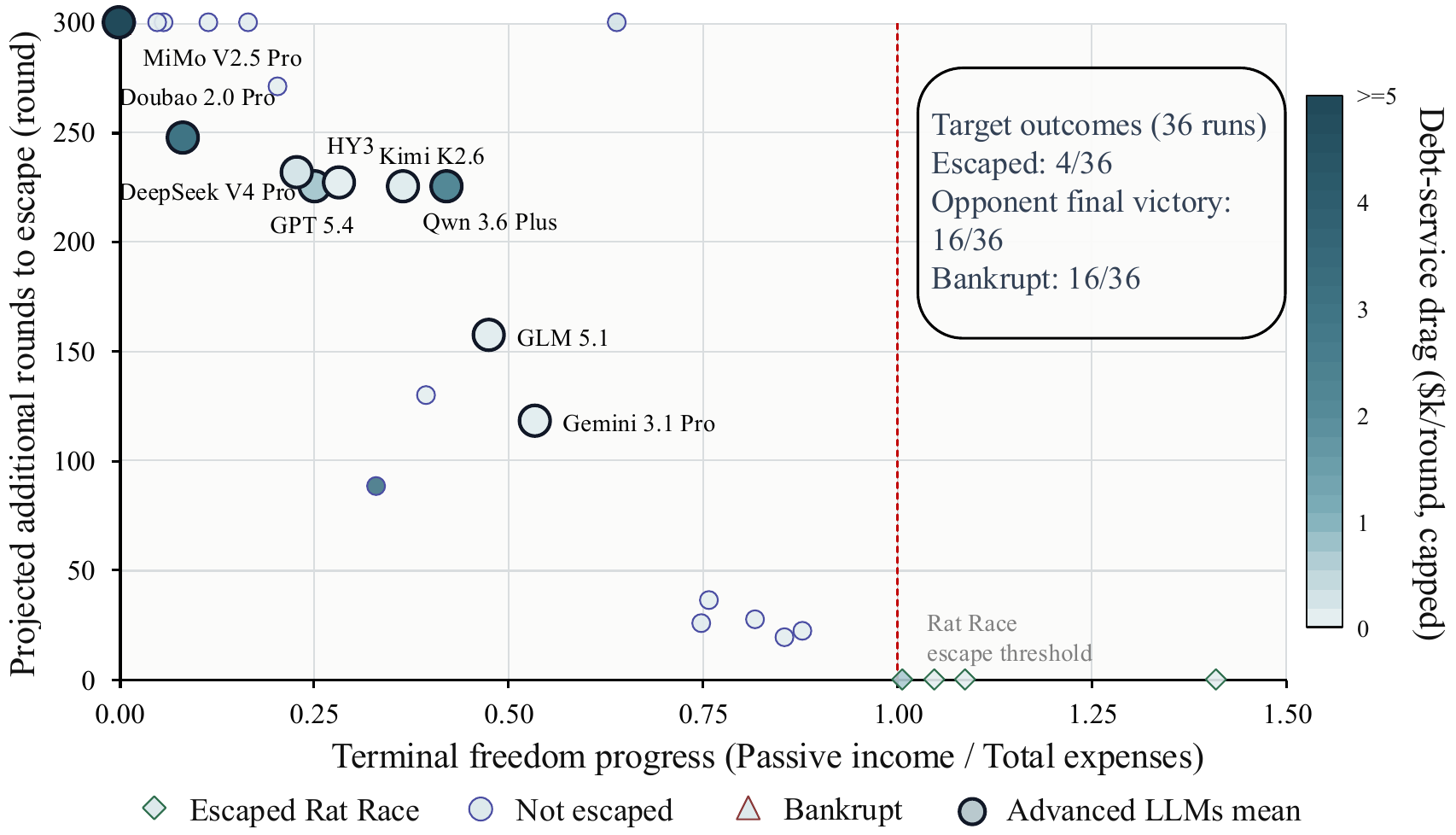}}
  \caption{Cashflow threshold-race results for Advanced LLMs. The x-axis shows terminal passive-income coverage of expenses, and the y-axis shows projected additional rounds to escape the Rat Race. Shapes denote outcomes, color denotes debt-service drag, and large outlined circles denote advanced LLM averages.}
  \label{Fig:cashflow_analysis}
\end{figure}

\begin{figure}[t]
  \centerline{\includegraphics[width=\columnwidth]{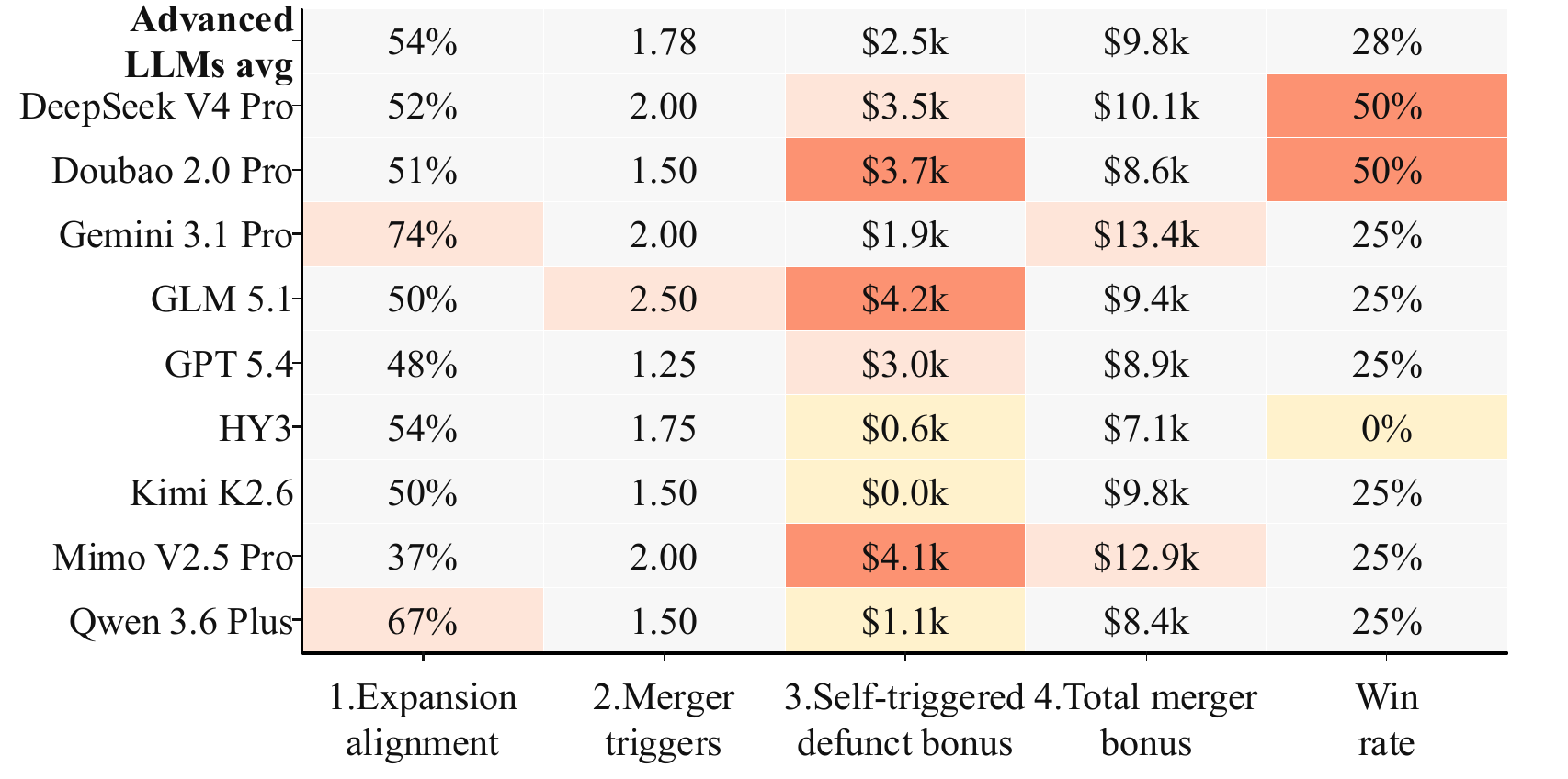}}
  \caption{Analysis of the behaviors of Advanced LLMs in Acquire. This heatmap depicts the complete acquisition arbitrage chain for Advanced LLMs: from ownership-aligned expansion to final victory. Each cell displays the raw value, while color encodes its value relative to the advanced LLM's average.}
  \label{Fig:acquire_analysis}
\end{figure}

\begin{figure}[t]
  \centerline{\includegraphics[width=\columnwidth]{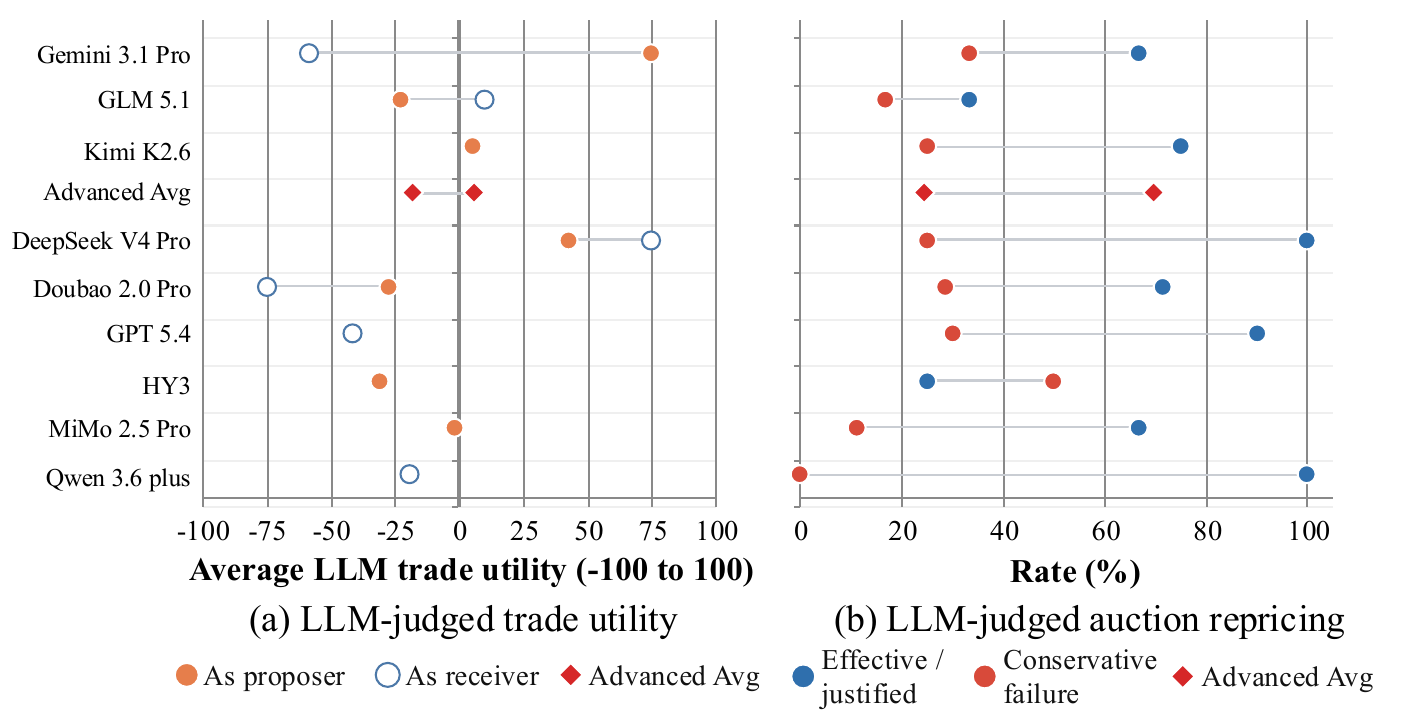}}
  \caption{Analysis of the trade and auction behaviors of Advanced LLMs in Monopoly.}
  \label{Fig:analysis_monopoly}
\end{figure}

\begin{figure}[t]
  \centerline{\includegraphics[width=\columnwidth]{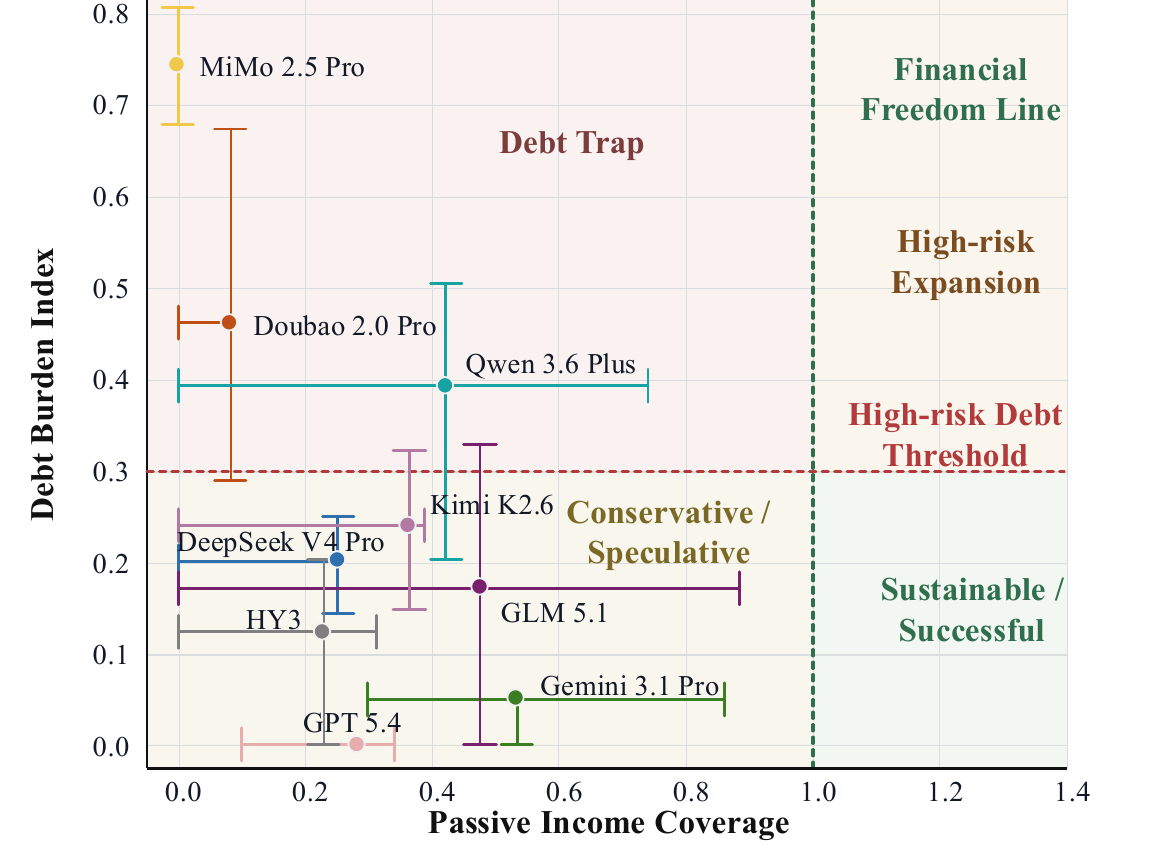}}
  \caption{Analysis of the financial statement profiles of LLMs in Cashflow. The x-axis represents the debt burden index (monthly loan payment / total expenses), and the y-axis represents the passive income coverage (passive income / total expenses).}
  \label{Fig:cashflow_profile}
\end{figure}

\subsection{Why do advanced LLMs fail?}

\textbf{LLMs fail to prioritize actions that accelerate financial freedom.}
As shown in Figure~\ref{Fig:cashflow_analysis}, advanced LLMs perform poorly in Cashflow, not only lack action, but also fail to convert actions into timely financial-freedom progress. The average performance of most LLMs is below the escape threshold. Furthermore, several non-bankrupt LLMs still require more than 100 projected rounds to escape. This suggests that they improve their financial situation too slowly in a the competitive and dynamic environment. In contrast, the darker points in the figure reveal that the debt servicing costs arising from borrowing disrupt the LLMs' path toward financial freedom, pushing it toward bankruptcy or a state of non-toward. In summary, success requires fast and stable conversion from asset expansion and financing decisions into sustainable cash flow progress.

\textbf{LLMs perform well in certain stages yet show poor overall performance.}
Success in Acquire depends on a sequential arbitrage chain: expanding chains where the LLMs hold control positions, triggering mergers, converting defunct-chain control into bonuses, and accumulating merger income toward victory. As shown in Figure~\ref{Fig:acquire_analysis}, advanced LLMs exhibit fragmented strengths across these stages. Gemini 3.1 Pro and Qwen 3.6 Plus perform well in controlled-chain expansion, with alignment rates of 74\% and 67\%, respectively. GLM 5.1 is strong at creating merger events, triggering 2.50 mergers per game and receiving \$4,200 in self-triggered defunct-chain bonuses. Mimo V2.5 Pro also captures high bonus income, with \$4,100 in self-triggered bonuses and \$12,900 in total merger bonuses. However, these advantages rarely transform into robust overall success. This indicates that advanced LLMs can perform isolated financial operations, but often fail to complete the full dynamic arbitrage chain required for consistent victory.

\textbf{LLMs cannot reliably convert trade and auction mechanisms into control.}
Trades are crucial for establishing property-group monopolies, while auctions reprice abandoned assets. We use DeepSeek V4 Pro as an external judge to evaluate each trade and auction from the full game state. Specifically, a trade is marked harmful if it reduces the focal LLMs' strategic or financial position, and an auction is marked conservative when the LLMs miss a valuable bidding opportunity, or the bidding increments are small. As shown in Figure~\ref{Fig:analysis_monopoly}(a), advanced LLMs often fail to convert successful trades into value: 45 of 86 proposer trades and 11 of 18 receiver trades are judged harmful. Figure~\ref{Fig:analysis_monopoly}(b) shows stronger auction behavior, with 42 of 60 cases judged effective or justified, but 15 still flagged as conservative failures. Overall, advanced LLMs can use trade and auction mechanisms, but lack the consistency needed to turn them into property control and a winning advantage.

\subsection{Decision Characteristics Analysis of Different LLMs}

The error bars in this section use quartiles to represent the volatility of the LLMs' performance. Specifically, they span the 25th-75th percentile range of the corresponding metric across multiple tests of the same LLM.

\textbf{Cashflow profiles reveal differences in financial management among advanced LLMs.} As shown in Figure~\ref{Fig:cashflow_profile}, Mimo V2.5 Pro, Doubao-2.0-Pro, and Qwen 3.6 Plus fall into the Debt Trap region, with low passive-income coverage coupled with high debt burden; Mimo V2.5 Pro is the most extreme case, with nearly zero coverage, a debt burden of 0.74, and a 100\% bankruptcy rate. In contrast, Gemini 3.1 Pro and GPT-5.4 maintain low debt burdens but still remain below the escape threshold, reflecting slow cash-flow conversion rather than over-leveraging. Other LLMs lie between these two distinct profiles. Overall, some LLMs over-expand through debt, while others avoid debt but fail to  generate passive income quickly enough.

\begin{figure}[t]
  \centerline{\includegraphics[width=\columnwidth]{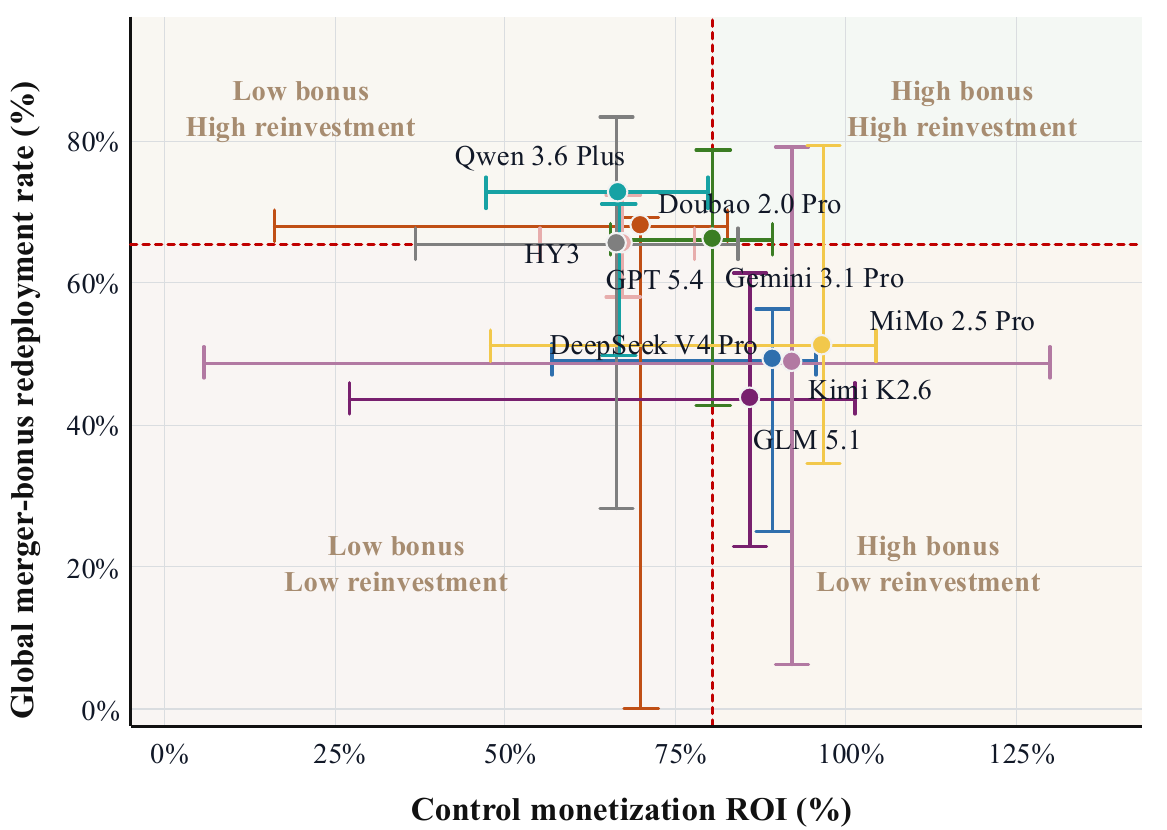}}
  \caption{Analysis of the decision profiles of LLMs in Acquire. The x-axis represents the control monetization ROI (merger bonus / stock purchase cost), and the y-axis represents the global merger-bonus redeployment rate (redeployed merger bonus / total merger bonus).}
  \label{Fig:acquire_profile}
\end{figure}

\begin{figure}[t]
  \centerline{\includegraphics[width=\columnwidth]{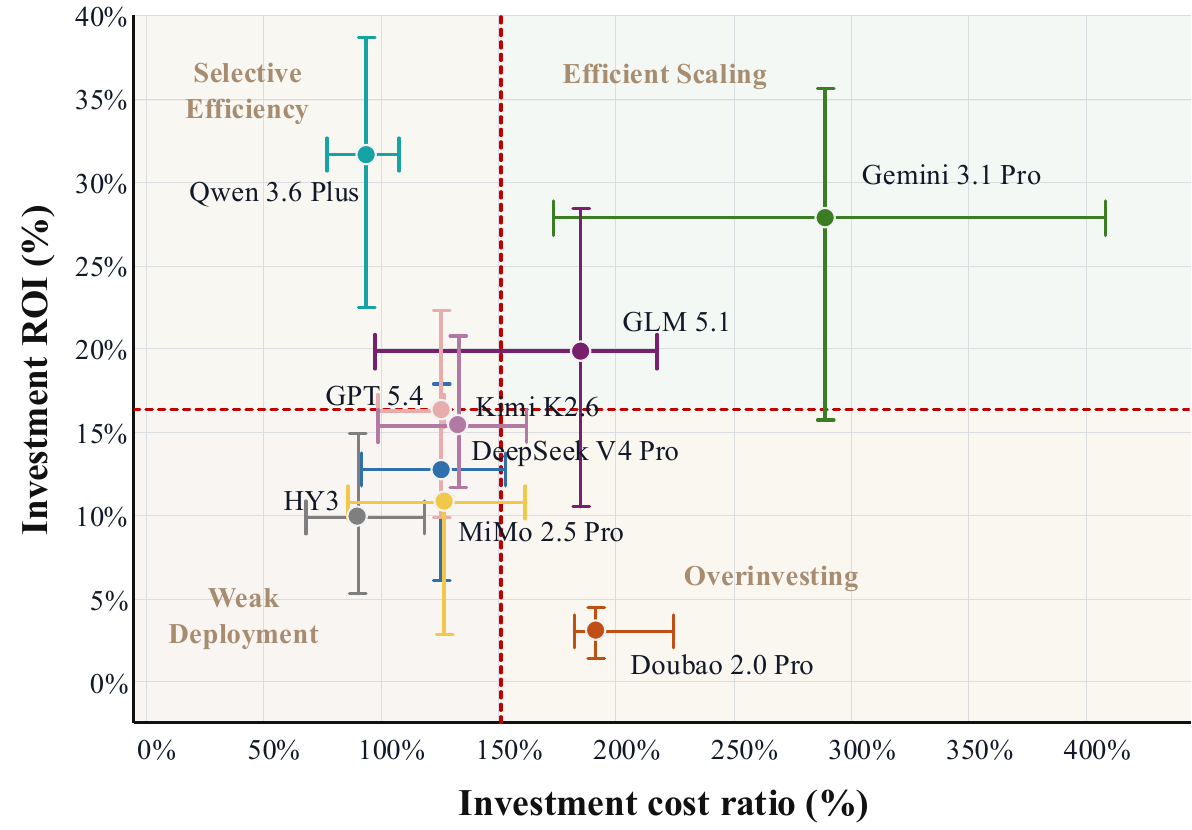}}
  \caption{Analysis of the investment profiles of LLMs in Monopoly. The x-axis represents the investment cost ratio (income / cost), and the y-axis represents the Investment ROI (total investment / initial cash).}
  \label{Fig:monopoly}
\end{figure}

\textbf{Acquire profiles reveal differences in decision-making among advanced LLMs.} As shown in Figure~\ref{Fig:acquire_profile}, advanced LLMs shows two different decision patterns. Mimo V2.5 Pro, Kimi K2.6, DeepSeek V4 Pro, and GLM 5.1 achieve higher Return on Investment (ROI) for control monetization. It indicates that their stock purchases more readily convert into merger bonuses, but their lower reinvestment rates limit subsequent deployment. In contrast, Qwen 3.6 Plus, Doubao 2.0 Pro, GPT 5.4, and HY3 reinvest more merger bonuses, yet demonstrate relatively lower efficiency in capturing control monetization returns. Gemini 3.1 Pro comes closest to the ideal state of simultaneously possessing strong monetization capabilities and reinvestment potential. Overall, Advanced LLMs usually master only one side of the Acquire arbitrage chain.

\textbf{Investment profiles reveal divergent decision strategies in Monopoly.} As shown in Figure~\ref{Fig:monopoly}, Gemini 3.1 Pro and GLM 5.1 fall into the Highly Efficient Scalable regime, both achieved above-average ROI while employing an aggressive investment strategy. Gemini 3.1 Pro is a particularly standout performer in this regard, with a win rate of up to 75\%. Doubao-2.0-Pro is situated in Overinvesting, investing heavily but yielding a low ROI. Qwen-3.6-Plus shows Selective Efficiency, while achieving high ROI with limited investment, it failed to transform this advantage into wins. Other models cluster in Weak Deployment, reflecting both low investment intensity and weak return conversion.

\section{Related work}

Many studies have proposed benchmarks for evaluating the capabilities of LLMs in the financial domain. FAMMA~\citep{xue2025famma} provides a multilingual, multimodal financial question answering benchmark for evaluating financial reasoning capabilities of LLMs. FINCHAIN~\citep{xie2026finchainsymbolicbenchmarkverifiable} is a verifiable Chain-of-Thought Financial reasoning benchmark spanning 12 Domains and 58 Topics. FinRule-Bench~\citep{malarkkan2026finrule} evaluates LLMs on structured financial auditing and multi-rule reasoning tasks.
Furthermore, some researchers are focused on dynamic simulations. StockAgent~\citep{zhang2024aimeetsfinancestockagent} simulates multi-agent trading to assess investor behavior under external influences. QuantAgent~\citep{xiong2025quantagent} combines multi-agent LLMs with structured trading signals to achieve higher prediction accuracy across high-frequency trading markets. MM-DREX~\citep{chen2025mm} uses multimodal-driven and dynamic expert routing, which outperforms baselines on multimodal datasets across stocks, futures, and cryptocurrencies. TwinMarket~\citep{NEURIPS2025_5bf234ec} reveals collective dynamics and emergent social behaviors by simulating socio-economic systems in which multiple LLMs interact.

Compared to these work, we are first to explore the financial decision-making capabilities of LLMs in dynamic games, and reveal the shortcomings of LLMs in this environment.

\section{Conclusion}
In this paper, to fully evaluate the dynamic financial decision-making capabilities of large language models, inspired by game-based benchmarks, we construct FinBoardBench. It is based on three classic financial board games and evaluates mainstream advanced LLMs. Experimental results demonstrate that while current LLMs perform well on static financial benchmarks, this capability does not transform into genuine financial decision-making ability, which leads to failures in LLMs. We hope FinBoardBench can serve as a valuable reference for future dynamic financial benchmarks.

\section*{Limitations}

First, FinBoardBench relies on board-game simulations, which may not fully capture the complexity of real-world finance, including exogenous shocks, interest-rate fluctuations, and similar factors. Second, using fixed game scenarios may limit evaluation under diverse or evolving financial conditions. For instance, FinBoardBench does not incorporate portfolios comprising agents with different strategic styles to evaluate LLMs under more dynamic multi-agent interactions. Finally, although we follow the official rules as closely as possible, the selected games are classic titles with extensive histories and multiple rule variants. Furthermore, to ensure compatibility for LLMs, we cannot reproduce every specific detail with absolute fidelity. Consequently, we focused on faithfully reproducing the core gameplay mechanics to the greatest extent possible.




\bibliography{custom}

\appendix

\section*{Appendix}

\section{Cashflow Game Rule}
\label{sec:cashflow_game_rule_appendix}
\subsection{Cashflow Game Background}
Cashflow is a board game designed to cultivate financial literacy. The board is divided into an inner and outer circle. Each player initially chooses a career and a dream. They start in the inner circle, rolling dice to move and handling the events on the spaces they land on. The goal here is to achieve passive income greater than total monthly expenses. Once the goal of the inner circle is achieved, players advance to the outer circle, where the goal becomes to further increase monthly cash flow to accumulate wealth or realize their dream.

\subsection{Game Map}
The game board contains two tracks: the Rat Race and the Fast Track. The Rat Race consists of an inner loop comprising 24 spaces. These spaces include Opportunity, Liability, Charity, Paycheck, Offer, Child, and Downsize. Opportunity spaces appear most frequently and serve as the main investment decision point. The Fast Track consists of an outer loop comprising 40 spaces. It contains Cashflow Day spaces, a Dream space, Charity, risk-event spaces, and business-investment spaces with explicit costs and monthly cash flow returns.

\subsection{Game Preparation}
Each player initially selects a career and a dream. The career determines the player's initial financial status, including salary, savings, and other variables. The dream defines the player's objective on the Fast Track. All players start in the Rat Race with their own income statement, balance sheet, and other relevant financial records. The performance of each player is evaluated through them. The key variables include cash, salary, passive income, total income, total expenses, monthly cash flow, assets, liabilities, and net worth. Passive income is derived from assets such as real estate, businesses, preferred stocks, certificates of deposit, limited partnerships, companies, coins, and contract receivables.

\subsection{Detailed Game Rules}
\textbf{Turn Structure in the Rat Race.} Each Rat Race turn consists of three major stages. First, the player may perform financial operations, including borrowing money or repaying liabilities. Specifically,
\begin{itemize}
  \item Bank loans must be at least \$1,000 and in multiples of \$1,000, with the monthly payment equal to 10\% of the borrowed amount.
  \item Loans may be repaid partially in \$1,000 units, while most other liabilities require full repayment. 
\end{itemize}

Second, the player rolls the dice and moves clockwise around the Rat Race. The events triggered by the landing spaces are as follows:

\begin{itemize}
  \item Opportunity cards represent investment decisions. Landing on an Opportunity space, the player chooses between the Small Deal and Big Deal card pools. Small Deal cards include stocks, mutual funds, certificates of deposit, preferred stocks, coins, small real-estate deals, and passive market events such as stock splits or property damage. Big Deal cards focus primarily on larger real-estate and business investments. Stock-like assets require the player to choose a purchase quantity and are paid in full. Real-estate and business assets are purchased through a down payment and, when specified, an associated mortgage or business liability. A successful purchase adds the asset to the player's balance sheet and updates passive income.
  \item Liability spaces draw Doodad cards and require the player to pay expenses or take on additional debt when applicable.
  \item Charity allows the player to donate 10\% of monthly total income to roll two dice for the next three turns.
  \item Paycheck spaces pay the player's monthly cash flow when passed or landed on. Bankruptcy is triggered when Paycheck processing leaves the player with negative cash and negative monthly cash flow. The engine then resolves bankruptcy through asset liquidation, debt repayment, debt relief, or elimination.
  \item Offer spaces trigger market offers for selling eligible assets. Depending on the card, the offer may be priced per unit, fixed per asset, or tied to a specific asset type. Selling an asset removes it from the player's balance sheet, repays related financing obligations, and credits the net settlement to cash. Some offers are global market events and may be resolved by all eligible players, not only by the player who landed on the Offer space.
  \item Child spaces increase monthly expenses, subject to a maximum of three children.
  \item Downsize forces the player to pay one month of expenses, cancels Charity effects, and skips two turns.
\end{itemize}

Third, after resolving the space event and completing available actions, the player may enter a second financial-operation phase before ending the turn. 

\subsection{Transition to the Fast Track}
A player immediately enters the Fast Track once passive income exceeds total expenses. Upon entry, the player receives a Buyout equal to 100 times their passive income. This amount becomes the player's Beginning CASHFLOW Day Income. The player's Fast Track winning target is then defined as Beginning CASHFLOW Day Income plus \$50,000. The fast track mechanics are as follows:

\begin{itemize}
  \item Fast Track players normally roll two dice. If Fast Track Charity has been unlocked, they may choose to roll one, two, or three dice each turn.
  \item Passing or landing on a Cashflow Day space grants the player's current Cash Flow Day Income.
  \item Fast Track investment spaces can be purchased with cash only; borrowing is not allowed. Purchasing Fast Track businesses increases Cash Flow Day Income and are removed from availability for other players. Some Fast Track opportunities are roll-based and only become claimed if the success roll is achieved.
  \item Fast Track risk spaces impose large cash or asset shocks, such as losing all cash, paying half of current cash, losing the lowest-cash-flow asset, or paying repair costs. 
  \item The Dream space allows the player to buy their dream if they have sufficient cash. 
\end{itemize}

\subsection{Winning Conditions}
A player wins Cashflow by satisfying one of two Fast Track victory conditions: purchasing their Dream, or increasing Fast Track cash flow by at least \$50,000 above the Beginning CASHFLOW Day Income.

\section{Acquire Rule}
\label{sec:acquire_game_rule_appendix}
\subsection{Acquire Game Background}
The Acquire game is a game through which you maximize your wealth by creating, merging, and acquiring hotel chains and buying their stock. Specifically, multiple players place tiles on the board. Two tiles form a chain, and players can then buy stock in the chain. When two chains connect, the smaller chain is merged into the larger chain. Players holding stock in the smaller chain can choose to sell, trade, or keep their stock. The larger the chain size, the higher the stock value. At the end of the game, all stock is converted into cash.

\subsection{Game Map}
The game map is a 12x9 game board. There are a total of 108 building tiles, 7 hotel chains, each hotel chain has 25 stocks, a total of 175 tickets. 

\subsection{Detailed Game Rules}
Players must buy and sell shares by observing the size and location of each chain, and considering potential merger opportunities to maximize bonuses and overall wealth. The relevant concepts in the game are as follows:

\textbf{Objective.} Players strive to accumulate the most cash by the end of the game. The primary sources of cash include stockholder bonuses during mergers and from selling stocks when the game ends. The player with the highest total cash is declared the winner.

\textbf{Key Concepts.} A hotel chain consists of two or more connected building tiles. Chains with 11 or more tiles are considered safe and cannot be acquired. The player who establishes a hotel chain is the founder and receives a founder's bonus stock. When a player places a building tile connecting two or more chains, a merger occurs: the smaller chain merges into the larger one, and stockholder bonuses are distributed. These bonuses are awarded exclusively to the top two controlling stockholders of the merged chain.

\textbf{Turn Sequence.} On each turn, a player performs three actions:

\begin{itemize}
  \item Place a building tile on its designated space, potentially founding a hotel chain, enlarging an existing chain, or triggering a merger.
  \item Optionally buy up to three stocks from any active hotel chains.
  \item Draw a new building tile from the pile. Mergers and corresponding stockholder bonuses are resolved immediately after tile placement.
\end{itemize}

\textbf{Stock Pricing and Bonuses.} Stock prices are determined by the specific hotel chain and its size. During mergers, players holding stock in acquired chains may sell, trade, or retain their holdings. Bonuses are distributed based on the ranking of stockholding dominance. In case of ties among stockholders, specific rules determine how bonuses are split.

\textbf{Endgame Conditions.} The game concludes when either all active hotel chains are safe or a hotel chain reaches 41 or more tiles. At this point, stock is converted to cash, all remaining bonuses are fully distributed, and the player with the highest cash total wins.

\section{Monopoly Game Rule}
\label{sec:monopoly_game_rule_appendix}
\subsection{Monopoly Game Background}
Monopoly is a multiplayer board game where players move around the board by rolling two dice. When you land on an unowned property, you can buy it and also trade properties with others. All properties of the same color must be collected to construct houses or hotels on them. When other players land on your property, they need to pay rent. The more buildings you have, the higher the rent will be. Every time you pass or land on the "GO" space, you collect \$200. The game also includes Chance and Community Chest cards. Drawing these cards can result in a gain or loss of money, or move you to another space. Some spaces require you to pay taxes. Players can also be sent to Jail and must meet specific conditions to get out. The goal is to drive all other players into bankruptcy.

\subsection{Game Preparation}
The board has 40 spaces, comprising 28 assets: 22 streets of 8 colors, 4 railroads, and 2 utility companies. There are also 3 chance spaces, 3 community spaces, 1 luxury tax space, and 1 income tax space. The four corners represent the starting point "GO", the Jail/Just Visiting, the Free Parking, and Go to Jail. All properties, houses, and hotels are held by the bank before the player purchases them. Each asset comes with a property deed specifying its price, mortgage value, construction costs for houses and hotels, and rent for different property grades. Specifically:

\begin{itemize}
  \item Cards: There are 32 cards, divided into 16 chance cards and 16 community chest cards.
  \item Railways: Rent increases with the number of railroads owned: 1 railroad: \$25, 2 railroads: \$50, 3 railroads: \$100, 4 railroads: \$200. Houses cannot be built on them.
  \item Streets: All properties of the same color must be collected before a house can be built. Buildings must be evenly distributed within the properties. Houses must be built on the properties with the fewest houses. When selling buildings, hotels must first be downgraded to 4 buildings, while houses are sold in grades. Banks repurchase properties at half price.
  \item Utilities (Water Works and Electric Company): Rent is a multiple of the dice roll. Owning a utility costs four times the dice's value; owning two utilities costs ten times the dice's value. Houses cannot be built on them.
  \item The four corner spaces: Starting at "GO" grants \$200; "Jail/Just Visiting" space has no effect when passed; "Free Parking" space does not trigger any events; "Go to Jail" space will directly send the player to prison.
\end{itemize}

\subsection{Detailed Game Rules}
\textbf{Game Initialization.}
Each player starts with \$1500 and rolls the dice once, determining the starting order based on the number rolled. Players move clockwise, rolling two dice each turn and moving the corresponding number of spaces along the board. If doubles are rolled in the same turn, the player gains an extra turn.

\textbf{Jail Rules.}
The following situations cause the player to enter the jail in the "Jail/Just Visiting" space and directly end the turn. Specifically: (1) Landing on the "Go to jail" space; (2) Rolling three consecutive doubles in a round; (3) Drawing the "Go Directly to Jail" card. Rules after entering jail are as follows:
\begin{itemize}
  \item Upon entering jail, the player cannot collect the \$200 at the starting point.
  \item The player cannot move while in jail. To leave jail, the player take follow actions: (1) Paying the \$50 fine; (2) Using the "Get Out of Jail Free" card; (3) Rolling a double. If the player fails to roll a double for three consecutive rounds, they must either pay the fine or use the "Get Out of Jail Free" card to leave.
  \item The player in jail can not purchase properties. They can conduct trade, upgrade buildings, collect rent, etc.
  \item If a player starts their turn in jail, they can choose to pay the fine or use the "Get Out of Jail Free" card to exit immediately, and then roll the dice normally to move. Rolling doubles grants an extra turn. If they choose not to pay or use a card, rolling doubles allows the player to exit and move, but without an extra turn. If they do not roll doubles, the movement phase is skipped.
\end{itemize}

\textbf{Property Management Rules.}
The player can purchase it from the bank at the listed price when landing on any unowned property. If the player declines the purchase, the bank should auction the property publicly, and all players can bid. Nothing happens when they reach their own property. If the players land on an unmortgaged property owned by others, they must pay the rent. If lacking sufficient cash, they must mortgage their own properties or sell houses to raise funds. Otherwise, the players go bankrupt. If a player owns all properties of the same color group, the rent for vacant properties in that group is doubled. They may then build houses and later hotels to increase the rent. Houses must be built evenly within a group; no property can have more than one house extra compared to any other property in that same group. The specific rules for houses and hotels are as follows:

\begin{itemize}
  \item Players can purchase houses from the bank at their round. After building four homes on the same property, the player can pay the cash to replace them with a hotel. 
  \item On a property, each house represents a level of land, with a hotel representing level 5. Homes and hotels can be sold back to the bank at half the purchase price.
  \item Unbuilt properties can be freely traded between players at agreed negotiations.
\end{itemize}

\textbf{Mortgage Rules.}
A player can mortgage property that has no buildings on it. The bank should pay them half of the property's purchase price. To unmortgage the mortgage, the player must repay this amount plus 10\% interest. While mortgaged, a property cannot collect rent. Mortgaged properties can be traded, but the player acquiring the property must pay the mortgage plus 10\% interest to make the property normal. Only properties without buildings can be mortgaged. If there are buildings on the property being mortgaged, all buildings must be sold before mortgaging the property.

\textbf{Bankruptcy Rules.}
When a player owes debts to other players or the bank, that player must be able to raise enough cash to pay off the entire debt. Players unable to repay their debts are considered bankrupt and eliminated; if a bankrupt player owes the bank, they must return all their property to the bank, which then demolishes all buildings and auctions them off; if the debt is owed to another player, all property is given to that opponent, except for buildings that must be sold to the bank.

\begin{itemize}
  \item If the new owner chooses to retain the property, they must repay any mortgages held by the bank on the received property plus 10\% interest.
  \item When a player is bankrupt and has no specific creditors, the bank repossesses all their assets.
\end{itemize}

\textbf{Winning Conditions.}
The winner is determined by the total assets of the players remaining after other players have gone bankrupt, including cash, the value of owned properties and buildings calculated at the purchase price, mortgaged properties calculated at half the purchase price, and hotels calculated at the purchase price plus the value of returned houses.

\section{Prompt Template}
We use a structured prompt for all FinBoardBench games. Each prompt consists of a stable system role, a current-state snapshot, phase-specific legal actions, and a strict JSON output standard. After each action is executed, the result is fed back into subsequent prompts, allowing the agent to update its belief state regarding assets, cash, liabilities, board positions, market events, auctions, and negotiations.

\begin{PromptBox}{FinBoardBench General Prompt}
\textbf{System role.}
You are playing a financial board game. You must follow the rules of the current game and use only the legal actions available in the current phase.

\textbf{Prompt construction.}
Each decision prompt contains:
\begin{itemize}
    \item \textbf{Static instruction:} game objective, core financial mechanics, and related constraints.
    \item \textbf{Dynamic state:} current player status, public board state, other players' public information, and recent history.
    \item \textbf{Phase context:} the active decision phase and the event that triggered it.
    \item \textbf{Action rules:} legal actions and parameter constraints for the current phase.
    \item \textbf{Output schema:} one strict JSON object.
\end{itemize}

\textbf{Common output rule.}
Return valid JSON only. Do not invent actions. Copy identifiers such as property IDs, chain names, asset IDs, liability IDs, and card names exactly from the prompt.
\end{PromptBox}

\begin{PromptBox}{Monopoly Prompt Template-P1}
\textbf{System role.}
You are playing Monopoly. Based on the current game state, decide your next action.

\textbf{State snapshot.}
The prompt provides color-group analysis, complete map overview, players overview, previous-round actions, property ID reference, player cash, owned assets, position, jail status, jail-card holdings, current space information, auction status, fund-raising status, and available houses/hotels.

\textbf{Normal-turn actions.}
\begin{itemize}
    \item \texttt{BUY}: buy the unowned property currently occupied.
    \item \texttt{UPGRADE[space\_id]}: upgrade one eligible owned property.
    \item \texttt{MORTGAGE[space\_id1,...]}: mortgage eligible undeveloped properties.
    \item \texttt{UNMORTGAGE[space\_id1,...]}: repay mortgage plus interest.
    \item \texttt{SELL\_BUILDING[space\_id]}: sell one building while preserving even-building constraints.
    \item \texttt{TRADE}: propose a trade of cash, properties, and jail cards.
    \item \texttt{USE\_JAIL\_CARD[card\_type]}, \texttt{PAY\_JAIL\_FINE}, or \texttt{END\_TURN}.
\end{itemize}

\textbf{Output format.}
For non-trade actions:
\texttt{\{"action": "ACTION\_TYPE[parameters]", "reason": "..."\}}.
For trade actions, output a JSON object containing the target player, offered/requested assets, offered/requested cash, offered/requested cards, and reason.
\end{PromptBox}

\begin{PromptBox}{Monopoly Managers Prompt Template-P2}
\textbf{Jail manager.}
The prompt provides jail turns, cash, jail cards, map context, player overview, and previous actions. Legal actions are \texttt{USE\_JAIL\_CARD[chance]}, \texttt{USE\_JAIL\_CARD[community\_chest]}, \texttt{PAY\_JAIL\_FINE}, and \texttt{END\_TURN}.

\textbf{Fund-raising manager.}
When cash is insufficient, the prompt provides the amount needed, shortage reason, the current cash, assets, mortgage options, sell-building options, failed attempts, and global context. Legal actions focus on liquidity recovery, such as \texttt{MORTGAGE[...]}, \texttt{SELL\_BUILDING[...]}, or bankruptcy when recovery is impossible.

\textbf{Auction manager.}
The prompt provides auction property, price, mortgage status, color group, rent, current bid, minimum bid, current bidder, auction round, eligible bidders, player cash, assets, and bid history. Legal actions are \texttt{BID[amount]} and \texttt{PASS}.

\textbf{Trade managers.}
The request manager decides whether to enter negotiation: \texttt{ACCEPT} or \texttt{REJECT}. The negotiation manager supports \texttt{ACCEPT}, \texttt{REJECT}, and \texttt{COUNTER}; counteroffers must use only owned tradable assets and cannot include properties with buildings.
\end{PromptBox}

\begin{PromptBox}{Acquire Prompt Template}
\textbf{System role.}
You are playing the Acquire game. You must follow the game rules strictly and decide your next action based on the current game state. Your core goal is to maximize your total wealth by strategically building hotel chains and investing in their stocks.

\textbf{State snapshot.}
The prompt provides money, stock holdings, hand tiles, legal tile placements, board state, active chains, other players' public information, remaining tiles, pending action, merger state, and end-game declaration eligibility.

\textbf{Phase decisions.}
\begin{itemize}
    \item \texttt{PLAY\_TILE}: output \texttt{\{"chosen\_tile": "XX", "reason": "..."\}}.
    \item \texttt{FOUND\_CHAIN}: output \texttt{\{"chosen\_chain": "CHAIN", "reason": "..."\}}.
    \item \texttt{CHOOSE\_MERGER\_SURVIVOR}: output \texttt{\{"surviving\_chain": "CHAIN", "reason": "..."\}}.
    \item \texttt{STOCK\_DISPOSITION}: output \texttt{\{"sell": n, "trade": n, "keep": n, "reason": "..."\}}.
    \item \texttt{BUY\_STOCKS}: output purchase list or \texttt{end\_turn}.
    \item \texttt{DECLARE\_END\_GAME}: output \texttt{\{"declare\_end\_game": true/false, "reason": "..."\}}.
\end{itemize}

\textbf{Key constraints.}
Safe chains cannot be merged; defunct-share trade counts must be even; stock purchases are limited by cash, availability, and turn quota.
\end{PromptBox}

\begin{PromptBox}{Cashflow Prompt Template-P1: Rat Race}
\textbf{System role.}
You are playing Cashflow. In the Rat Race, your goal is to build Passive Income greater than Total Expenses so you can leave the Rat Race and enter the Fast Track.

\textbf{Financial mechanics.}
Cash is immediate liquidity. Total Income equals Salary plus Passive Income. Monthly Cash Flow equals Total Income minus Total Expenses. Bank loans increase cash but add monthly payments equal to 10\% of the borrowed principal. Bankruptcy is checked at Pay Check when cash and the Monthly Cash Flow remain negative.

\textbf{State fields.}
The prompt provides player name, career, dream, cash, position, salary, expenses, passive income, assets, liabilities, charity turns, current card or offer, recent events, failed attempts, and situation analysis.

\textbf{Rat Race phases.}
\begin{itemize}
    \item Career/Dream setup: \texttt{CHOOSE\_CAREER}, \texttt{CHOOSE\_DREAM}.
    \item \texttt{FINANCIAL\_OP}: \texttt{REPAY\_LIABILITY}, \texttt{BORROW\_MONEY}, \texttt{END\_FINANCIAL\_OPERATION}.
    \item \texttt{OPPORTUNITY\_CHOICE}: \texttt{CHOOSE\_SMALL\_DEAL} or \texttt{CHOOSE\_BIG\_DEAL}.
    \item \texttt{REGULAR}: \texttt{BUY\_OPPORTUNITY}, \texttt{SELL\_ASSET}, \texttt{ACCEPT\_OFFER}, \texttt{DONATE\_CHARITY}, \texttt{END\_TURN}.
    \item \texttt{PURCHASE\_LOAN\_OFFER}, \texttt{GLOBAL\_OFFER}, and \texttt{BANKRUPTCY\_RESOLUTION}.
\end{itemize}

\textbf{Output format.}
Return \texttt{\{"action": "EXACT\_ACTION", "parameters": \{...\}, "reason": "..."\}}.
\end{PromptBox}

\begin{PromptBox}{Cashflow Prompt Template-P2: Fast Track}
\textbf{Fast Track transition.}
In the Fast Track, the goal changes to winning by buying your Dream or reaching the required Fast Track cash-flow goal.

\textbf{Objective.}
The player wins by buying the Dream or increasing Current Cash Flow Day Income by at least \$50,000 above Beginning Cash Flow Day Income.

\textbf{State fields.}
The prompt provides current cash, Beginning Cash Flow Day Income, Current Cash Flow Day Income, additional cash flow gained, remaining win gap, Fast Track position, current square, Fast Track assets, complete Fast Track map, player overview, and charity status.

\textbf{Fast Track phases.}
\begin{itemize}
    \item Dice choice: \texttt{CHOOSE\_FT\_DICE\_COUNT} with dice count 1, 2, or 3 when charity is unlocked.
    \item Action phase: \texttt{BUY\_FT\_OPPORTUNITY}, \texttt{DONATE\_FT\_CHARITY}, \texttt{BUY\_FT\_DREAM}, or \texttt{END\_FT\_TURN}.
\end{itemize}

\textbf{Output format.}
Return \texttt{\{"action": "EXACT\_FAST\_TRACK\_ACTION", "parameters": \{...\}, "reason": "..."\}}. Borrowing and Rat Race debt actions are not allowed on the Fast Track.
\end{PromptBox}

\section{Fault Tolerance Mechanisms}
To ensure the stable operation of the game, we have set decision validation, error feedback, and system takeover mechanisms. Every decision generated by the LLMs undergoes format parsing, action validity checks, and execution result verification. When a format, parameter, or execution error occurs, the environment feeds back the specific error type and the concrete reason to the model, thereby enabling it to make a revised decision within the same situational context. Furthermore, fault tolerance mechanisms exist with retry limits. Such as in Monopoly, the main decision-making phase permits a maximum of 5 action execution failures. Once the retry limit is reached, the environment initiates a conservative, system-enforced fallback, such as ending the turn, advancing to the next stage, and selecting a default valid option. These interventions are explicitly recorded as system-forced events to distinguish them from decisions made autonomously by the model.

\end{document}

%% file: table/intro_.tex
\begin{table}[t]
    \centering
    \begin{adjustbox}{width=1.000\columnwidth}
    \begin{tabular}{lccccc}
      \toprule
      Financial Benchmark & Competitive & \makecell[c]{Multi-\\Rounds} & \makecell[c]{Comprehensive\\Strategy} & Uncertain \\
      \midrule
      PIXIU~\citep{xie2023pixiu} & \errormark & \errormark & \errormark & \errormark \\
      FinBen~\citep{NEURIPS2024_adb1d9fa} & \errormark & \errormark & \errormark & \errormark \\
      FinMME~\citep{luo-etal-2025-finmme} & \errormark & \errormark & \errormark & \errormark \\
      BizFinBench~\citep{lu2025bizfinbench} & \errormark & \errormark & \errormark & \errormark \\
      FinMaster~\citep{jiang2025finmaster} & \errormark & \errormark & \errormark & \errormark \\
      SECQUE~\citep{benyoash-etal-2025-secque} & \errormark & \errormark & \errormark & \errormark \\
      StockBench~\citep{chen2025stockbench} & \errormark & \correctmark & \errormark & \correctmark \\
      Investorbench~\citep{li-etal-2025-investorbench} & \errormark & \correctmark & \errormark & \correctmark \\
      \midrule
      \textbf{FinBoardBench} (Ours) & \correctmark & \correctmark & \correctmark & \correctmark \\
      \bottomrule
    \end{tabular}%
    \end{adjustbox}
    \caption{Comparison of FinBoardBench with previous financial benchmarks. FinBoardBench provides multi-round competitive interactions under uncertainty through comprehensive financial games.}
    \label{tab:compare}%
\end{table}%